\PassOptionsToPackage{colorlinks=purple,
  linkcolor=purple,
  citecolor=purple,
  urlcolor=purple,
  pdfauthor={},
  pdftitle={},
  pdfsubject={},
  pdfkeywords={},
  bookmarks=false,}{hyperref}
\documentclass[sigplan,nonacm, 10pt, screen]{acmart}

\usepackage[usenames,dvipsnames,table]{xcolor}
\usepackage{hyperref}

\setcopyright{acmlicensed}
\copyrightyear{2018}
\acmYear{2018}
\acmDOI{XXXXXXX.XXXXXXX}

\acmConference[Conference acronym 'XX]{Make sure to enter the correct
  conference title from your rights confirmation emai}{June 03--05,
  2018}{Woodstock, NY}
  \acmISBN{978-1-4503-XXXX-X/18/06}

\usepackage{amsmath,amsfonts}
\usepackage{graphicx}
\usepackage{tikz}
\usepackage{nomencl}
\usepackage{amsthm}
\usepackage{diagbox}
\usepackage{soul}
\usepackage{multirow}
\usepackage{textcomp}
\usepackage[normalem]{ulem}
\usepackage{caption}
\usepackage{subcaption}
\usepackage{multicol}

\usepackage[nopostdot,nonumberlist,acronyms]{glossaries}

\usepackage[utf8]{inputenc}
\usepackage[english]{babel}
\usepackage{pifont}

\newcommand{\cmark}{{\color{OliveGreen} \ding{51}}}
\newcommand{\xmark}{\color{BrickRed} \ding{55}}

\usepackage[noend]{algpseudocode}
\usepackage{algorithm}

\definecolor{Maroon}{RGB}{176, 48, 96}

\usepackage{float}
\usepackage{caption}
\usepackage{subcaption}
\usepackage{graphicx}

\usepackage{multirow}
\usepackage{diagbox}

\usepackage[normalem]{ulem}
\usepackage{textcomp}
\usepackage{xspace}

\usepackage{flushend}

\usepackage[draft,commandnameprefix=ifneeded,truncate=hyphenate]{changes}

\definecolor{cgreen}{rgb}{0.0, 0.42, 0.24}

\graphicspath{{figures/}}

\definecolor{orange}{RGB}{225,128,0}
\definecolor{brown}{RGB}{225,128,128}
\xdefinecolor{DarkGreen}{cmyk}{0.8,0,0.8,0.3}
\definecolor{rgviolet}{RGB}{119,51,255}
\definecolor{redviolet}{RGB}{255,0,85}
\definecolor{jdbrown}{RGB}{77,38,0}

\definecolor{orange}{RGB}{225,128,0}
\definecolor{brown}{RGB}{225,128,128}
\definecolor{purple}{rgb}{0.54, 0.17, 0.89}
\definecolor{rggold}{RGB}{218,165,32}
\definecolor{rgviolet}{RGB}{119,51,255}

\newcommand{\ma}{Spyker\xspace}

\newcommand{\ms}{Sync-Spyker\xspace}

\makeglossaries

\newacronym{iid}{IID}{independent and identically distributed}
\newacronym{tee}{TEE}{Trusted Execution Environment}
\newacronym{lstm}{LSTM}{Long Short-Term Memory}
\newacronym{kde}{KDE}{Kernel Density Estimate}

\begin{document}

\def\sectionautorefname{Sec.}
\def\subsectionautorefname{Sec.}
\def\figureautorefname{Fig.}
\def\tableautorefname{Tab.}
\def\algorithmautorefname{Alg.}
\def\equationautorefname{Eq.}

\title{Asynchronous Multi-Server Federated Learning for Geo-Distributed Clients}

\author{Yuncong Zuo}
\email{Y.Zuo-2@student.tudelft.nl}
\affiliation{
  \institution{Delft University of Technology}
  \city{Delft}
  \state{}
  \country{Netherlands}
}
\renewcommand{\shortauthors}{Zuo et al.}
\author{Bart Cox}
\email{b.a.cox@tudelft.nl}
\orcid{0000-0001-5209-6161}
\affiliation{
  \institution{Delft University of Technology}
  \city{Delft}
  \state{}
  \country{Netherlands}
}

\author{Lydia Y. Chen}
\email{lydiaychen@ieee.org}
\orcid{0000-0002-4228-6735}
\affiliation{
  \institution{Delft University of Technology}
  \city{Delft}
  \state{}
  \country{Netherlands}
}

\author{J\'{e}r\'{e}mie Decouchant}
\email{j.decouchant@tudelft.nl}
\orcid{0000-0001-9143-3984}
\affiliation{
  \institution{Delft University of Technology}
  \city{Delft}
  \state{}
  \country{Netherlands}
}

\begin{abstract}
Federated learning (FL) systems enable multiple clients to train a machine learning model iteratively through synchronously exchanging the intermediate model weights with a single server.  The scalability of such FL systems can be limited by two factors: server idle time due to synchronous communication and the risk of a single server becoming the bottleneck. In this paper, we propose a new FL architecture, to our knowledge, the first multi-server FL system that is entirely asynchronous, and therefore addresses these two limitations simultaneously.  
Our solution keeps both servers and clients continuously active. As in previous multi-server methods, clients interact solely with their nearest server, ensuring efficient update integration into the model. Differently, however, servers also periodically update each other asynchronously, and never postpone interactions with clients.
We compare our solution to three representative baselines -- FedAvg, FedAsync and HierFAVG -- on the MNIST and CIFAR-10 image classification datasets and on the WikiText-2 language modeling dataset.  
Our solution converges to similar or higher accuracy levels than previous baselines and requires 61\% less time to do so in geo-distributed settings.

\end{abstract}

\settopmatter{printfolios=true}
\maketitle

\section{Introduction}
\label{sec:introduction}

Federated Learning (FL) is the emerging paradigm to train machine learning models iteratively on extensive data that are dispersed across many users and cannot be openly shared due to privacy concerns or regulations~\cite{fl_proposed}. The FL paradigm is widely adopted in applications such as mobile keyboard prediction~\cite{edge2}, spoken language comprehension~\cite{app1}, and digital healthcare~\cite{app2,health0, health1,health2}.
The most prevalent FL architecture is composed of a single-server and multiple clients, and assumes a synchronous  training process over multiple rounds~\cite{fl_proposed}. In each round, the server selects a set of clients that receive the current global model, train it with their local dataset, and send a model update back to the server. Once it has received all model updates, the server aggregates them and computes the new global model that is used in the subsequent round. The training process is repeated until the global model converges.

\begin{figure}[t]
    \centering
    \includegraphics[width=\columnwidth]{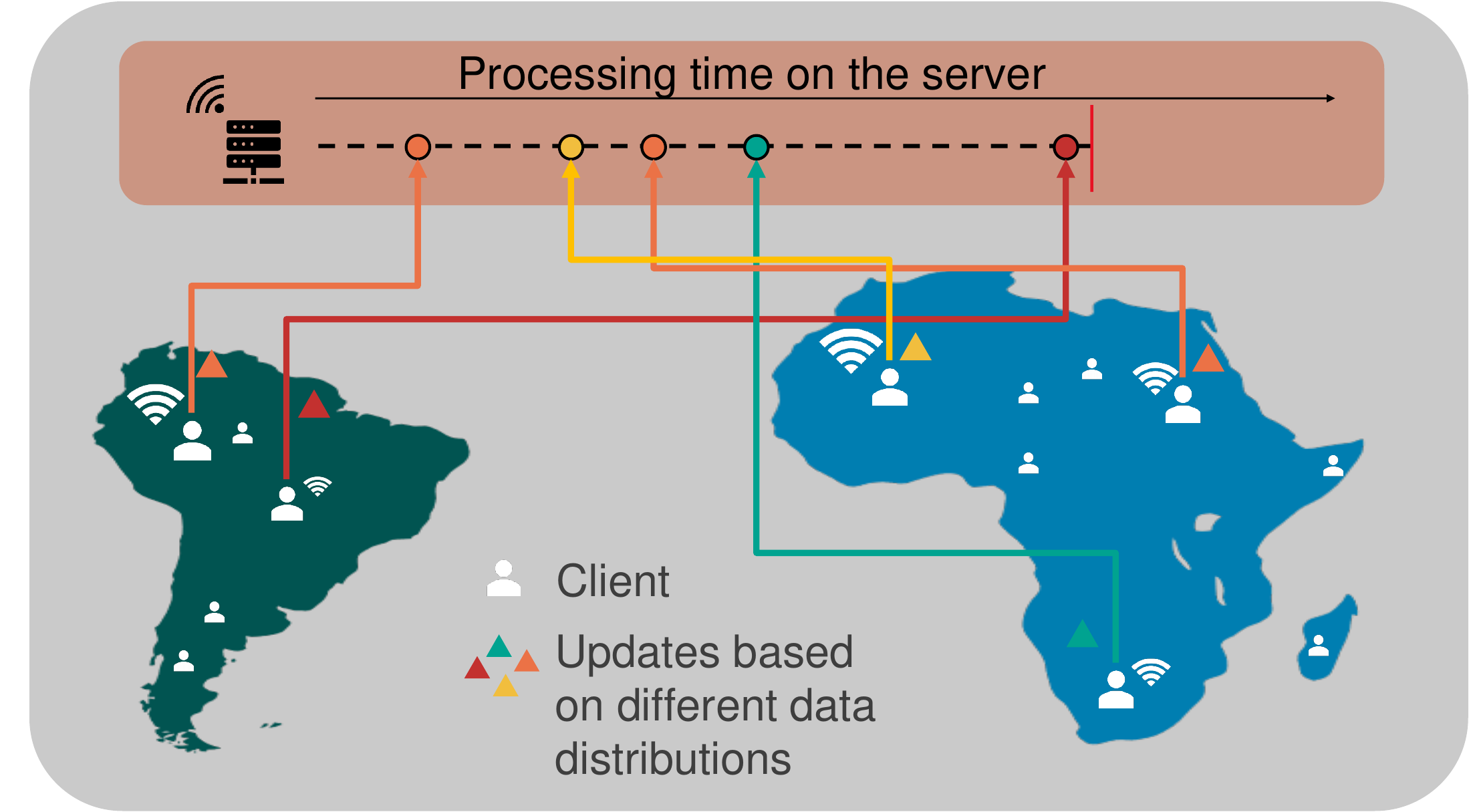}
    \caption{A training round of a synchronous FL system with heterogeneous devices and networks. The times at which the server receives client updates are indicated with colored spheres. The server waits for all client updates to be received to update the model, which results in a mostly idle server.} 
    \label{fig:fl_frame}
\end{figure}

\begin{table*}[t]
	\centering
	\caption{Comparison of representative multi-server FL algorithms with \ma.}
	\label{tab:comparison_table}
	\rowcolors{2}{gray!10}{white}
		\begin{tabular}{|l|c|c|c|c}
            \hline
			\rowcolor{gray!25}  & {\bf Mutli-Server} & {\bf Async Client/Server} & {\bf Async Server/Server}  \\
            \hline
            FedAvg~\cite{fl_proposed} & \xmark & \xmark & N/A\\
            FedAsync~\cite{xie2019fedasync} & \xmark & \cmark & N/A\\
            HierFAVG~\cite{hier_sync} & \cmark & \cmark & \xmark \\
			\hline
           {\bf \ma} & \cmark & \cmark & \cmark \\ 
           {\bf \ms} & \cmark & \cmark & \xmark \\ 
           \hline      
		\end{tabular}
\end{table*}

The duration of a training round in a synchronous FL framework is determined by the longest client update time, because a round only completes when all updates of  selected clients are received. 
Clients exhibiting heterogeneous computational capacity and network bandwidth can unfortunately prolong the update time, degrading the efficiency of FL systems.
The computational heterogeneity stems from the difference in their hardware and software stacks; whereas  network latency difference is rooted in the geo-distributed settings where clients are spread across several countries or continents~\cite{cano2016towards,hsieh2017gaia,zhu2022sky}. The result of such heterogeneity is often system inefficiency: the server, along with clients near it or those with superior computing power, remain underutilized. 
\autoref{fig:fl_frame} illustrates this problem with a synchronous FL round that involves five clients distributed over two continents. The server has to collect all five client updates before updating the global model. The first four clients become idle after they send an update to the server and the server remains idle until the end of the round. 

Asynchronous FL frameworks~\cite{xie2019fedasync, wang2022asyn, asyn_fl} aim at eliminating performance bottlenecks arising from heterogeneity in client's computation and networking. Under such frameworks, the server needs to immediately process a client's model update upon receiving it and return the update global model to the corresponding client.
Thus, slow clients in an asynchronous FL framework no longer hamper the overall performance. However, this can inadvertently shift the bottleneck to the server. Specifically, if the server is continuously busy processing client updates, it might cause waiting periods for clients, in particular as the number of clients increases. 

To prevent potential server-based bottlenecks, one can deploy multiple servers that each interact with a subset of the clients. In particular, hierarchical FL frameworks~\cite{client_edge_cloud, hier_sync, smartPC}  rely on secondary servers that interact with clients, and on a primary server that merges the models of the secondary servers. This multi-server structure decreases the average communication latency between a client and a server, given that clients upload their model updates to the nearest server. However, in such hierarchical FL frameworks, the primary server still relies on a synchronous procedure to update its global model, which in geo-distributed settings implies that secondary servers stop replying to clients. 

We introduce \ma\footnote{The \ma was the first four-wheel-drive car. The wheels of such a car are similar to our FL framework's servers: they make progress asynchronously and independently from each other, and occasionally synchronize.}, a novel multi-server and fully asynchronous FL framework that relies on a flat architecture of geo-distributed servers. 
Moreover, interactions between clients and their nearest server, and between servers themselves are asynchronous to maintain low response times, and reduce both client and server idle times. \ma accelerates the training of a model in geo-distributed settings, where heterogeneous clients and servers are located in different geographical regions.    

Table~\ref{tab:comparison_table} compares \ma to the relevant related work and highlight its main characteristics

\textbf{Contributions.} This work makes the following contributions to realize the asynchronous multi-server FL vision:

\begin{itemize}
\item 
To support multi-server asynchronous FL, we define the age of a server model and the staleness of a client update.  Based on these definitions, a server determines the weight that server models or client updates should be given when aggregating them into its local model.    

\item To maintain high accuracy despite resource heterogeneity, 
\ma triggers an exchange and aggregation of server updates when the ages of two server models differ too much or when the age of a model is sufficiently changed since the last server model aggregation. To minimize the complexity of the server model aggregation process, in particular in asynchronous networks, a single server at a time can trigger the exchange of server models. This decision is helped by a token that servers circulate among themselves and that collects the age of all server models. 

\item To further maintain accuracy despite data heterogeneity, i.e., clients having non-independent and identically distributed (non-iid) datasets, we detail a local learning rate decay strategy that prevents server models from becoming biased towards the data distributions of fast clients.  

\item We evaluate \ma's performance and compare it to three recent and representative FL frameworks, namely FedAvg~\cite{fl_proposed}, HierFAVG~\cite{hier_sync} and FedAsync~\cite{xie2019fedasync}. We also consider a variant of \ma, which we call \ms, where servers use a synchronous model exchange protocol. We evaluate all frameworks in emulated geo-distributed settings using the MNIST and CIFAR-10 image datasets and the WikiText2 language modeling dataset. We report the accuracy and convergence speed of all frameworks, and evaluate their ability to scale with the numbers of clients and servers. 
\end{itemize}

Our results show that \ma converges to similar or higher accuracy levels than previous baselines with a 61\% shorter running time in geo-distributed settings. \ma also scales better than other baselines with the number of clients thanks to its flat multi-server architecture. For example, increasing the number of clients from 100 to 200 multiplies the time our three baselines require to reach 90\% accuracy by at least 1.64, while \ma's convergence time is only multiplied by 1.21 (i.e., scalability is improved by 26\%). 

This paper is organized as follows. 
\autoref{sec:background} provides some background on FL systems and introduces the most commonly used synchronous, asynchronous, and hierarchical FL algorithms.
\autoref{sec:system_model} describes our system model and gives an overview of \ma. 
\autoref{sec:multiasync} discusses \ma in detail, including the local update and global model exchange strategies. 
\autoref{sec:performance_evaluation} presents the results of our performance evaluation.
\autoref{sec:related_work} discusses the related work, and~\autoref{sec:conclusion} concludes this paper.

\section{Background}
\label{sec:background}

\begin{table}[b]
    \caption{Notations and parameter values}
    \label{tab:notations} 
    \vspace{-3mm}
    \begin{center}
        \resizebox{\columnwidth}{!}{
            \begin{tabular}{llp{5cm}}
                Symbol & & Description \\
                \hline
                
                $D_k$ & \vspace{0mm} & Local dataset of client $k$\\
                $d_k$ & & Number of data points in $D_k$\\
                $d$ & & Total number of data points $\sum_k d_k$ \\
                
                $n$ & & Number of servers \\
                $n_C$ & & Number of clients \\
                $W^t_k$ & & Weight vector of a model at client $k$  \\ 
                
                $\eta_k$ & & Learning rate of client/server $k$ \\

                $T_k$ & & Number of local epochs on client $k$ \\
                $A_k$ & & Model age of client/server $k$ \\
                
                \hline
                \hline
                Symbol & Value & Description \\
                \hline
                $\eta_{init}$ & 0.5 & Initial client learning rate \\
                $\eta_{min}$ & $10^{-6}$ &  Minimum client learning rate \\
                $\beta$ & 0.05 & Decaying rate  \\
                $h_{\mathrm{inter}}$ & $n_C/5n$ & Age drift threshold between server models \\
                $h_{\mathrm{intra}}$ & 350 & Age drift threshold since last global aggregation\\
                $\phi$ & 1.5 & Activation rate \\
                $\eta_{\alpha}$ & 0.6 &  Aggregation rate (server side) \\
            \end{tabular}
        }
    \end{center}
\end{table}

\begin{figure*}[t]
    \centering
    \begin{subfigure}[b]{0.32\textwidth}
        \centering
        \includegraphics[width=.8\textwidth]{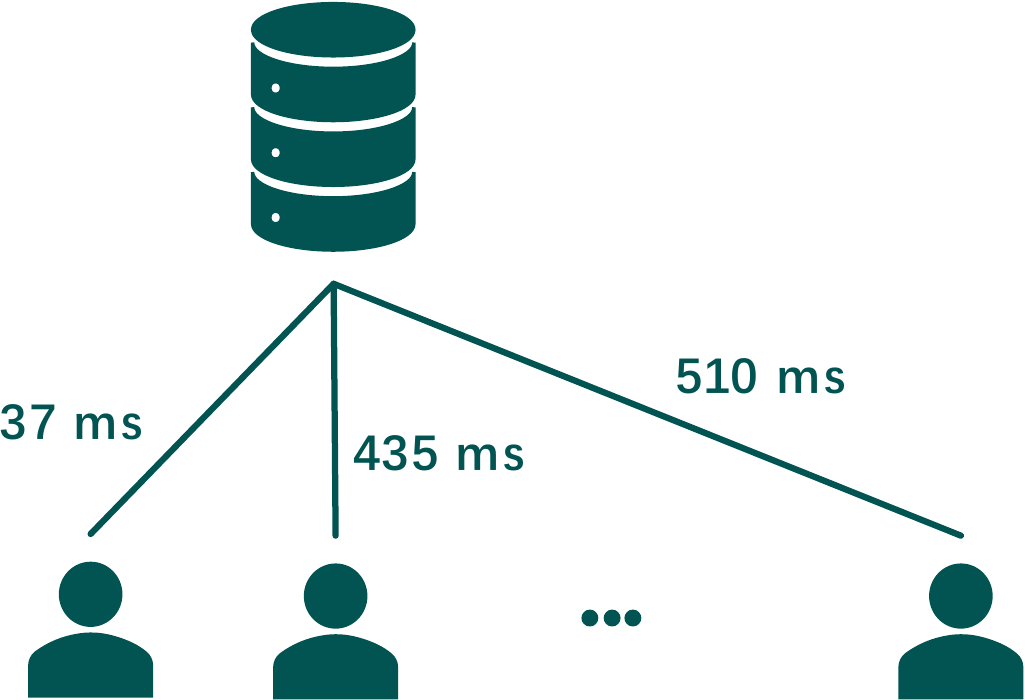}
        \caption{Centralized}
        \label{fig:centralized}
    \end{subfigure}\hfill 
    \begin{subfigure}[b]{0.32\textwidth}
        \centering
        \includegraphics[width=\textwidth]{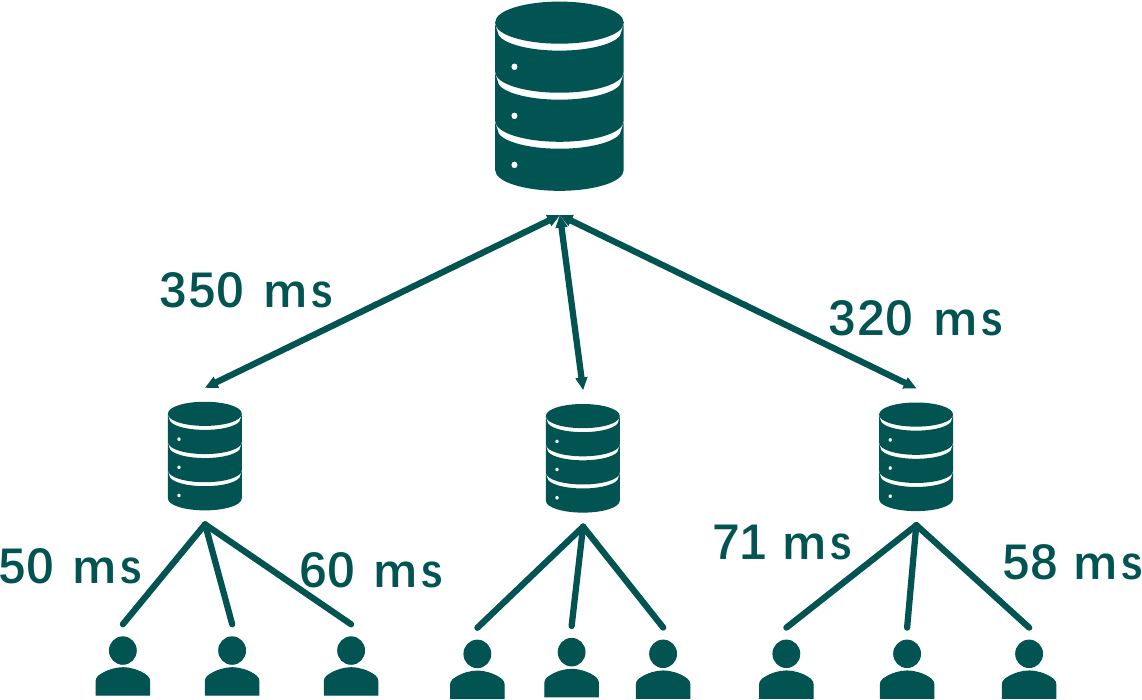}
        \caption{Hierarchical Multi-Server}
        \label{fig:hierarchical}
    \end{subfigure} \hfill
    \begin{subfigure}[b]{0.32\textwidth}
        \centering
        \includegraphics[width=\textwidth]{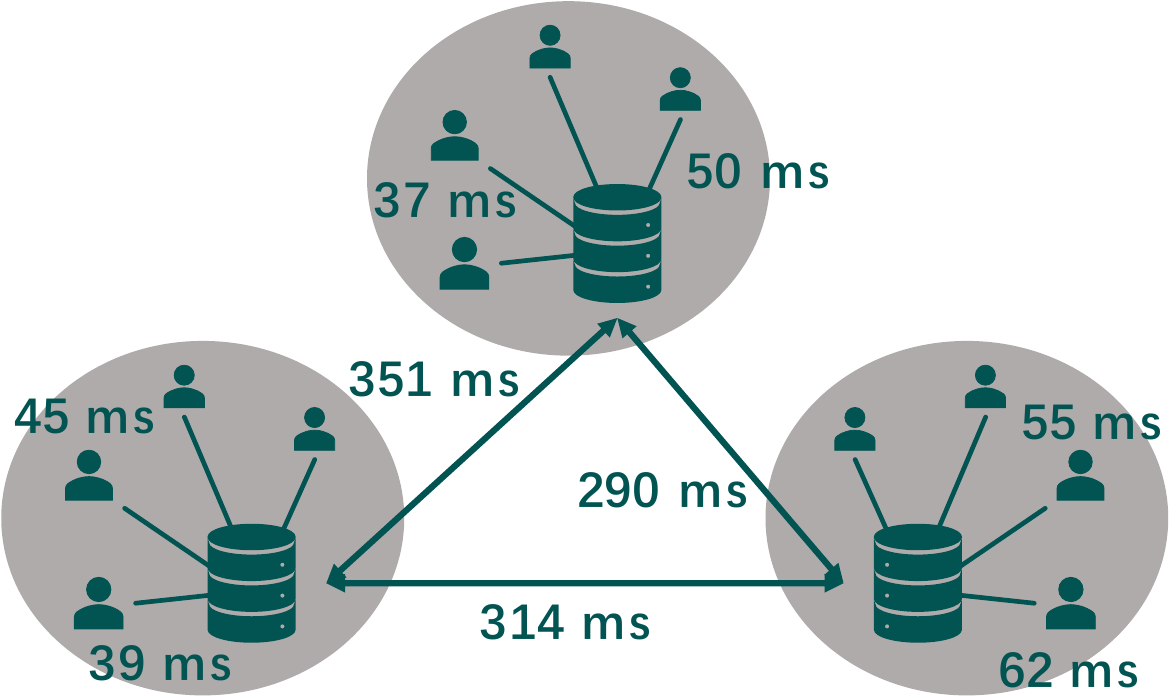}
        \caption{Flat Multi-Server (used in this work)}
        \label{fig:decentralized}
    \end{subfigure}
    \caption{Architectures of FL systems}
    \label{fig:system_models}
\end{figure*}

This section provides necessary background on synchronous, asynchronous and multi-server Federated Learning. \autoref{fig:system_models} illustrates the architectures these paradigms rely on, and \autoref{tab:notations} summarizes the notations we use throughout the paper.   

\subsection{Synchronous Federated Learning}
\label{subsec:synchronous_federated_learning}

The first FL framework was proposed by McMahan et al. in 2017~\cite{fl_proposed}. It relies on a classical centralized architecture where a server interacts with all clients (\autoref{fig:centralized}).  
The first FL algorithms, namely FedAvg and FedSgd, are synchronous. In each round, the server selects a group of clients and sends them the latest global model. Upon receiving a global model $W^t$, a client $k$ trains it on its local dataset $D_k$ to obtain $W_k^{t+1}$ using~\autoref{eq:fl_1}, where $F$ is the loss function of the classification, and $\eta_k$ is the learning rate. 

\begin{equation} \label{eq:fl_1}
    W_k^{t+1} = W_k^{t} - \eta_k \frac{\partial F(W^t, D_k)}{\partial W^{t}},
\end{equation}

Afterwards, client $k$ sends its model update $W_k^{t+1}$ to the server. Upon receiving all client updates it expects in a given round, the server aggregates them to compute the next global model $W^{t+1}$ using~\autoref{eq:fl_2}. In this equation, $d_k$ is the number of data points of client $k$ and $d$ is the total number of data points across all clients.

\begin{equation} \label{eq:fl_2}
    W^{t+1} = \sum_{k=1}^{K}{\frac{d_k}{d} W_{k}^{t}}.
\end{equation}

The ratio $\frac{d_k}{d}$ is used to weight the participation of a client during aggregation. At the beginning of the following round, the server sends $W^{t+1}$ to the next selected clients. The training process is repeated over multiple rounds until the global model converges. In practical settings, where clients have different computing speeds and are connected to the server with heterogeneous network links, synchronous FL algorithms may lead to the server and some clients staying idle. For example, one client in~\autoref{fig:centralized} would communicate with the server within only 37\,ms. This client and the server would have to stay idle waiting for the clients with 435\,ms and 510\,ms communication latency with the server to send their update.  

\subsection{Asynchronous Federated Learning}
\label{subsec:asynchronous_federated_learning}

Asynchronous FL systems, such as FedAsync~\cite{xie2019fedasync}, accelerate convergence in heterogeneous networks and under client heterogeneity. To do so, the server aggregates an update into a global model immediately after it receives it, and returns the new global model to the client. 
For example, a server using the asynchronous version of FedAvg in classical centralized settings (\autoref{fig:centralized}) would update the global model using~\autoref{eq:afl}. In this equation, $W^{t}$ is the $t$-th global model, $W^{k}_{t}$ is the local update client $k$ computed on $W^{t}$, $\frac{d_k}{d}$ is the data proportion of the update, and $s(\tau)$ is a staleness parameter that dampens the effect of client $k$'s update if it relates to a model that is older than the one the server currently has. 

\begin{equation} \label{eq:afl}
    W^{t+1} = W^{t} - s(\tau)\frac{d_k}{d} \left(W^{t}_{k}-W^{t+1}_{k}\right)
\end{equation}

An asynchronous FL scheme aims at keeping the clients busy training a model on their datasets, and the server updating its model more frequently, both of which are expected to speed up the model convergence. However, asynchronous schemes also increase the computational and communication loads on the server, which might become a performance bottleneck if the number of clients is too high. 

\subsection{Multi-server Federated Learning}
\label{subsec:multiserver_federated_learning}

Multi-server FL systems rely on multiple servers that interact with disjoint subsets of the clients to limit their computational and communication loads.  
Hierarchical FL systems (\autoref{fig:hierarchical}), such as HierFAVG~\cite{hier_sync}, are multi-server and rely on a principal server to maintain the global model~\cite{hier_sync, client_edge_cloud, HierFedAsync, smartPC}. In those systems, a model is updated in two times in each round. The first aggregation is managed by the edge servers that aggregate model updates from a group of clients and send their own model update to the principal server. Upon receiving all model updates from the servers, the principal server executes the second-level aggregation and generates the global model for the new round. The global model is then distributed to all servers, and eventually to all clients. 
Previous hierarchical FL frameworks are synchronous. In geo-distributed settings, where the servers would be located far from each other, a round in a hierarchical FL framework involves several successive long distance communications and model aggregations, which would limit its speed. 

\section{Overview of \ma}
\label{sec:system_model}

In this section, we start the presentation of \ma by providing the key ideas behind its design. \\

\textbf{Flat multi-server architecture.}
To get the best of both the asynchronous and hierarchical FL paradigms, \ma relies on multiple servers that directly interact with each other and with clients in a fully asynchronous manner. 
The servers are organized in a flat multi-server infrastructure (\autoref{fig:decentralized}). Each server is the center of a star sub-network, and interacts with a group of clients that are assigned to it based on geographical proximity to minimize communication latency. 
In this way, communication latency between clients and servers are reduced, and servers are always able to process client or server updates. On the contrary, the related studies mainly consider that clients interact with a single server. \\

\vspace{5mm}
\textbf{Asynchronous training and communications.}
Clients interact with their assigned server in the same way they would in a classical asynchronous FL system: they receive a global model from the server, train it over their local dataset and send their model update back to the server. 
\ma's servers rely on asynchronous communications to exchange their models and  benefit from the dataset of their respective clients. To minimize the complexity of this procedure, servers rely on a token-based strategy to trigger the asynchronous exchange of their models. Model aggregation also happens asynchronously and in a peer-to-peer fashion among servers. The model exchange can only be initiated by the server that holds the token, which prevents potential concurrent and redundant model broadcasts.  At a given point in time, servers might maintain slightly different models. \\

\textbf{Handling model staleness and fast clients.}
Since all interactions in \ma are asynchronous, servers take into account model staleness at all levels to maintain accurate models. To do so, servers maintain the age of their model. A server increases the age of its model whenever it processes a client model update, or the model of another server. \ma uses the age of models to dampen the contribution of an old model update when processing it. Similarly, \ma uses a learning rate decay to limit the influence of clients that frequently produce updates. \\

\textbf{Preventing server models from drifting apart.}
The frequency at which servers synchronize their models significantly influences the performance of the system in presence of data and resource heterogeneity. It is therefore important to ensure that a server model does not become biased toward the datasets of its clients, and that it leverages all client datasets. 
The model exchange and aggregation procedures allow servers to asynchronously homogenize their models. To prevent server models from drifting too much from each other during the training process, servers exchange and aggregate their models whenever they detect that their ages differ too much, and whenever a server has updated its model too many times since the last model exchange. 

\section{\ma Details}
\label{sec:multiasync}

In this section, we present the details of \ma's building blocks. 
We first detail how servers and clients interact. We then discuss \ma's aggregation algorithm that merges two server models and the token-based algorithm that triggers the asynchronous exchange of server models. \autoref{tab:notations} details our notations  and the values of all parameters. 

\subsection{Local Training and Decaying Learning Rates}
\label{subsec:client_server_aggregation}

\autoref{alg:local_update} describes the asynchronous interactions between a client and a server in \ma, which are divided into two main procedures.

The first procedure, \textsc{LocalTraining}, is executed by a client $C_k$ to train a model over its local dataset (\autoref{alg:local_update}, ll.~\ref{list:line:local_training}-\ref{list:line:send_w}), and is triggered upon receiving a model $W_i^t$, along with its age $A_i$ and a learning rate $\eta_k$ from a server. The client then trains the model over its dataset using the specified learning rate (\autoref{alg:local_update}, ll.~\ref{l:begin_training}-\ref{l:end_training}), and sends the trained model back to the server along with the age $A_i$ (\autoref{alg:local_update}, l.~\ref{list:line:send_w}). Note that instead of having the client sends the model's age with its update, the server could remember the age of the model it has sent to every client. We adopted this option to simplify our pseudocode. 

The second procedure, \textsc{Aggregation}, is executed by a server upon receiving a client update (\autoref{alg:local_update}, ll.~\ref{list:line:aggr}-\ref{list:line:send_w_to_client}). The server first determines the age difference between its current model and the model it sent to the client (\autoref{alg:local_update}, l.~\ref{l:age_diff}), and uses it to possibly decrease the impact of the received update on its model when updating it (\autoref{alg:local_update}, l.~\ref{l:aggr}). The server then computes the learning rate $\eta_k$ that the client should use  for its next local training (\autoref{alg:local_update}, l.~\ref{list:line:decay}), before returning its new model $W_i^{t+1}$ along with $A_i$ and $\eta_k$ to client $C_k$ (\autoref{alg:local_update}, l.~\ref{list:line:send_w_to_client}). Finally, the server verifies whether it should trigger a synchronization of server models (\autoref{alg:local_update}, l.~\ref{line:check_synchro}) if its model age has sufficiently increased since the latest synchronization by calling function $\textsc{checkSynchronization}()$ (defined at l.~\ref{line:checkSynchronization} in Alg.~\ref{alg:token_based}).  

\begin{algorithm}[t]
    \caption{Interactions between clients and servers \label{alg:cs}}
    \label{alg:local_update}
    \begin{algorithmic}[1]

        \item[$\eta_i$: server $S_i$'s learning rate] 
        \item[$\eta_k$: client $C_k$'s learning rate (given by $S_i$)]
        
        \item[]
        \renewcommand{\algorithmicprocedure}{\textbf{[Client $C_k$] Procedure}}
        \Procedure{LocalTraining}{$W^{t}_i, A_i, \eta_k$}\label{list:line:local_training}
            \State \Comment{receive model $W^{t}_i$ with age $A_i$ from server $S_i$}
            \State $W_k^{t} = W^{t}_i$

            \For{\textbf{each} epoch $ \in T_k $} \label{l:begin_training}
                \State update $W_k$ with learning rate $\eta_{k}$ \label{l:end_training}
            \EndFor
            \State $W_k^{t+1} = W_k^{t}$
            \State send $\left(W_k^{t+1}, A_i\right)$ to server $S_i$\label{list:line:send_w}
        \EndProcedure
        
        \item[]
        \renewcommand{\algorithmicprocedure}{\textbf{[Server $S_i$] Procedure}}
        \Procedure{Aggregation}{$W_k^{t'}, A_k$}\label{list:line:aggr}
            \State \Comment{receive model $W_k^{t'}$ with age $A_k{=}t'$ from client $C_k$}
            \State $w_k^t = A_i - A_k$ \label{l:age_diff}
            
            \State $W^{t+1}_i = W^{t}_i + \eta_i \cdot w^t_k \cdot \left(W_k^{t'} - W_i^{t}\right)$ \label{l:aggr}
            \State $A_i = A_i + 1$
            \State $u[k] = u[k] + 1$ \Comment{num. updates received per client} \label{l:incr_u}
            \State $\eta[k] = \mathrm{Decay}(\eta[u[k]], u[k], \overline{u})$ \label{list:line:decay}
            \State send $\left(W^{t+1}_i, A_i, \eta[k] \right)$ to client $C_k$ 
            \label{list:line:send_w_to_client}
            \State $\textsc{checkSynchronization}()$ \label{line:check_synchro}
        \EndProcedure
    \end{algorithmic}
\end{algorithm}

Let us provide more information on the way a server in \ma{} updates the  learning rate of a client (\autoref{alg:local_update}, l.~\ref{list:line:decay}) to maintain high model accuracy despite clients and network heterogeneity, i.e., despite the fact that some fast clients might consistently send updates at short intervals, while slow clients might contribute infrequently. 
\ma adopts an adaptive strategy that tailors the impact of each client update. More precisely, a server uses function $\mathrm{Decay}$ (\autoref{alg:local_update}, l.~\ref{list:line:decay}) to decrease the impact of the fast clients that frequently generate updates on its model. To do so, the server maintains the number of updates it has received from each client in an array $u$, and therefore increments $u[k]$ (\autoref{alg:local_update}, l.~\ref{l:incr_u}) upon receiving a model from $C_k$.  
The server adjusts client $C_k$'s learning rate using the following $\mathrm{Decay}$ function: 

\begin{equation*} \label{eq:decay}    
        \begin{tabular}{cc}
        $\mathrm{Decay}(\eta[u[k]], u[k], \overline{u}) = $ & \\
        $\left \{
        \begin{aligned} 
            & \eta_k\ & \mathrm{if}\ u[k] < \overline{u} \\
            & \max{\left(\eta_{min}, \eta\left[u[k]\right] - \beta(u[k] - \overline{u})\right)} & \mathrm{if}\  u[k] \geq \overline{u}  
        \end{aligned} \right.$
        & \\
        \end{tabular}
\end{equation*}

The $\mathrm{Decay}$ function takes three inputs: $u[k]$ is the number of updates that the server received from client $C_k$; $\eta[u[k]]$ is the learning rate a client would use without decay, which classically decreases with $u[k]$;  and $\overline{u}$ is the average number of updates that clients have sent to server $S_i$. $\mathrm{Decay}$ also uses two parameters: $\eta_{min}$ is a lower bound for learning rates; and $\beta$ is the decaying rate. We empirically determined that $\beta=0.05$ and $\eta_{min}=0.000001$ provide the best results (cf.~\autoref{tab:notations}).  

\subsection{Token-Triggered Server Model Aggregations}
\label{subsubsec:model_exchange_process_of_multiasync}

\begin{algorithm*}
    
    \caption{Token-triggered aggregation of server models.}\label{alg:cs}
    \label{alg:token_based}

    \begin{multicols}{2}
    \begin{algorithmic}[1] 
        
        \renewcommand{\algorithmicprocedure}
        {\textbf{[Server $S_i$]}} 
        \Procedure{ServerInit()}{} \label{line:server_init}
            \State $token = (i {\neq} 1)?\ NULL: \{bid{=}1, ages{=}[0, \cdots, 0]\}$
            \State $cnt = ages_i = \{\}$ \Comment{hashmaps}
            \State $didBroadcast = \emptyset$
            
            \State $ongoingSynchro = \mathrm{False}$
            \State $A_{i,prev} = A_i = 0$ 
            \State set $W_i^0$ to random model
        \EndProcedure

        \item[]
        \renewcommand{\algorithmicprocedure}{\textbf{[Server $S_i$] Procedure}}
        \Procedure{RcvAge}{$A_j$} 
             \label{line:rcv_age}
             \State ${ages}_i[j] = \mathrm{max}({ages}_i[j], A_j)$
            \State $\textsc{checkSynchronization}()$
        \EndProcedure
    
        \item[]
        \renewcommand{\algorithmicprocedure}{\textbf{[Server $S_i$] Procedure}} 
        \Procedure{RcvToken}{t} 
             \label{line:begin_rcv_token}
             \For{$j \in [1, N]$}{}
                \State ${ages}_i[j] = \mathrm{max}({ages}_i[j], t.ages[j])$
             \EndFor
             \State $token = t$
             \State $t.{bid} = t.{bid}+1$
             \State $\textsc{checkSynchronization}()$ \label{line:end_rcv_token}
        \EndProcedure

        \item[]
        \renewcommand{\algorithmicprocedure}{\textbf{[Server $S_i$] Procedure}} 
        \Procedure{checkSynchronization()}{} \label{line:checkSynchronization}
             \label{l:XXX}
                \If{$\mathrm{max}(ages) {-} \mathrm{min}(ages) {\ge} h_{\mathrm{inter}}$  \textbf{or} $A_i {-} A_{i,pre} {\ge} h_{\mathrm{intra}}$} 
                
                \If{$hasToken$ \textbf{and} (\textbf{not} $ongoingSynchro$)}
                    \State $A_{i,pre} = A_i$ 
                    \State $ongoingSynchro = \mathrm{True}$
                    \State send $(W_i^t, A_i, t.bid)$ to all servers  \label{line:broadcast_model}
                    \State $didBroadcast = didBroadcast \cup \{t.bid\}$
                    \State ${cnt}_i [token.bid] = 1$ 
                \Else 
                    \State send $A_i$ to all servers \label{line:broadcast_age}
                \EndIf

            \EndIf
        \EndProcedure

        \renewcommand{\algorithmicprocedure}{\textbf{[Server $S_j$] Procedure}} 
        \Procedure{RcvModel}{$W_i^t, A_i, {bid}_i$}\label{lst:line:rec_broadcast}
            \State ${ages}_j[i] = \mathrm{max}({ages}_j[i], A_i)$
  
                \If{\textbf{not} ${bid}_i \in didBroadcast$} 
                    \State $didBroadcast = didBroadcast \cup \{{bid}_i\}$
                    \State $A_{j, pre} = A_j$
                    \State send $\left(W_j^t, A_j, {bid}_i\right)$ to all servers
                \EndIf
                \State $\textsc{ServerAgg}(W_i^t, A_i)$ \label{lst:line:server_aggr}
                \If{$token \neq \mathrm{NULL}$ \textbf{and} $t.bid = {bid}_i$}   
                    \State $cnt[{bid}_i] = cnt[{bid}_i]+1$ \label{l:cnt}
                    \If{$cnt[{bid}_i] = n$} 
                        \State $t.ages = ages_j$ 
                        \State send token $t$ to the next server on ring \label{lst:line:pass_token_second}
                        \State $token = \textrm{NULL}$
                        \State $ongoingSynchro = \textrm{False}$
                    \EndIf
                \EndIf      
            
        \EndProcedure
        \item[]
        \renewcommand{\algorithmicprocedure}{\textbf{[Server $S_i$] Procedure}}
        
        \Procedure{ServerAgg}{$W_j^t, A_j$} \label{line:server_agg}
            \State process model $W_j^t$ of age $A_j=t$ from server $S_j$
            \State $a = \frac{\phi(A_j-A_i)}{A_i}$ \label{l:begin_weights} 
            \State $w_{i,j} = {\left(1+e^{- a}\right)}^{-1}$ \label{l:end_weights}
            \State $W_i^{t+1} = {W_i^t + \eta_a \cdot w_{i,j} \left(W_j^t-W_i^t \right)}$
            \State $A_i = \left(1-\eta_a \cdot w_{i,j}\right) A_i + \eta_a \cdot w_{i,j} \cdot A_j$ \label{l:age_update}
        \EndProcedure
    \end{algorithmic}

    \end{multicols}

\end{algorithm*}

In addition to interacting with their clients, \ma{}'s servers synchronize their models using \autoref{alg:token_based}. Since the age of a server model impacts the way its client updates are aggregated, \ma{} triggers an asynchronous server model synchronization whenever the ages of server models are drifting too much or when a server has processes a large enough number of client updates to make sure that all client datasets are fairly represented in server models. 

To avoid the complexity of having to handle multiple concurrent synchronizations, servers rely on a token so that only one server can trigger the model broadcasts and aggregations.   This token contains a synchronization ID $bid$ that allows server to broadcast their model only once per synchronization. Only the server holding the token can trigger a synchronization, but other servers might indirectly learn about an ongoing synchronization and broadcast their models to each other. A server broadcasts its model either because it holds the token and because some conditions are met (more details below), or by receiving another server's model with an unknown synchronization ID. To maintain fairness, the token is circulated among all servers. To accelerate synchronizations, whenever necessary, servers can broadcast their age so that the token holder triggers a synchronization. 
Note that we assume that links are FIFO, which can easily be enforced if it is not the case, so that a server receives the models of any other server according to increasing synchronization IDs and can process them all. 

Upon executing the \textsc{ServerInit} initialization procedure (\autoref{alg:token_based}, l.~\ref{line:server_init}), a server $S_i$ initializes several variables to coordinate the exchange of models between servers. The token initially resides at a randomly chosen server (here server $S_1$), and contains a synchronization ID $bid$ set to $1$ and a vector of server model ages that are all initially equal to $0$. The hashmap $cnt$ is used to count the number of models that have been received for a given synchronization ID. The known age of server models is kept in ${ages}_i$. 
Variable $didBroadcast$ contains the set of synchronization IDs for which $S_i$ has broadcast its model. The Boolean variable $ongoingSynchro$ is used by the server holding the token to initiate a synchronization only once before forwarding the token. The model age at which the latest synchronization happened, and the current model age are respectively stored in $A_{i,prev}$ and $A_i$. Finally, a server randomly initializes its model.  

The token keeps visiting every server following a random ring-based topology. Upon arrival of the token (\autoref{alg:token_based}, ll.~\ref{line:begin_rcv_token}--~\ref{line:end_rcv_token}), a server $S_i$ updates the ages of the models that it maintains using the token's information. It then calls the \textsc{CheckSynchronization} procedure (\autoref{alg:token_based}, l.~\ref{line:checkSynchronization}) to determine whether a model synchronization should be triggered. The frequency at which server model exchanges happen is controlled by the $h_{\mathrm{inter}}$ and $h_{\mathrm{intra}}$ parameters (cf.~\autoref{tab:notations}):  $h_{\mathrm{inter}}$ is the threshold for the maximum age difference between different server models; and $h_{\mathrm{intra}}$ is the threshold for the maximum age difference between a server model's current age and the age it had during the last server model synchronization. If the inter-server age difference exceeds threshold $h_{\mathrm{inter}}$, or if the intra-server age difference exceeds threshold $h_{\mathrm{intra}}$, then server $S_i$ might trigger a server model exchange. First, if $S_i$ holds the token and has not already triggered a server model exchange, then it broadcasts its current model $W_i^t$ of age $A_i$ along with the current broadcast ID $t.bid$ to all the other servers. Second, if $S_i$ does not hold the token, it can broadcast its model age to all other servers, so that the server that holds the token can execute procedure \textsc{RcvAge} (\autoref{alg:token_based}, l.~\ref{line:rcv_age}), which updates the age of a server model and possibly triggers a model synchronization. 

Procedure $\textsc{RcvModel}$ (\autoref{alg:token_based}, ll.~\ref{lst:line:rec_broadcast}-\ref{lst:line:pass_token_second}) is executed when a server receives a server model, and is also in charge of relaying the token. Upon receiving a server model from a server $S_i$, server $S_j$ first updates its local age for server model $i$. It then verifies whether it has already broadcast is model for the synchronization ID indicated by server $S_i$ using its local $didBroadcast$ set, and if necessary broadcasts its model and updates the relevant variables. A server that receives a server model then  executes the aggregation procedure $\textsc{ServerAgg}$ (\autoref{alg:token_based}, l.~\ref{line:server_agg}), which is described in details in the next section. Finally, the server that holds the token makes sure that it receives the model of all other servers (using the $cnt$ hashmap) before forwarding the token to its successor on the ring of servers.  

\subsection{Aggregation of Server Models}
\label{subsec:model_aggregation}

Following their exchange of models, servers aggregate them asynchronously immediately after they are received. \autoref{alg:cs} also details the procedures that a server $S_i$ follows to aggregate a model that it receives from another server $S_j$. This algorithm first calculates the weight that should be used to update the local server model with the received one.  
The aggregation weight $w_{i,j}$ is computed based on the age of the two server models $A_i$ and $A_j$ using a sigmoid function that is parameterized using parameter $\phi$ (\autoref{alg:cs}, ll.~\ref{l:begin_weights}-\ref{l:end_weights}): 

\begin{equation*}   \label{eq:age_weighted_aggregation}
    w_{i,j} = \frac{1}{1+e ^{-a}},\quad \text{where} \quad a = \frac{\phi(A_j-A_i)}{A_i}
\end{equation*}

The age difference $A_j-A_i$ indicates whether $W_j^t$ is more mature than $W_i^t$ in terms of the number of updates both models have been trained on. The influence of $W_j^t$ increases with this age difference. Denominator $A_i$ makes the absolute weight relative to the current age of server $S_i$. As a model age increases, it becomes more stable and the impact of other server models should be decreased. The sigmoid function ensures that $w_{i,j}$ remains between $0$ and $1$. The derivative of this function becomes $0$ and results in a weight of $1$ when the relative model age difference is too large. Parameter $\phi$ indicates in which range the sigmoid function is active. A larger $\phi$ leads to a smaller active range, leaving a larger area for the weight of 0 and 1.  

The aggregation rate $\eta_a$ also scales the influence of other servers during model aggregation. When the aggregation with server $S_i$ happens locally at server $S_j$, a large $\eta_a$ indicates that server $S_i$ could greatly influence server $S_j$'s model, and $\eta_a = 1$ means it could replace server $S_j$'s model when their ages are equal. Parameter $\eta_a$ needs to be carefully tuned. If $\eta_a$ is too large, the influence of the server itself is eliminated; if $\eta_a$ is too small, the server learns too little from its peers thus its model could be biased to its clients' data distribution. 
We empirically determined $\phi = 1.5$ and $\eta_\alpha = 0.6$ to provide the best results (cf.~\autoref{tab:notations}). 

Server $S_i$ finally updates its age after aggregating another server model. According to the weight $w_{i,j}$ that has been applied to the model aggregation, server $S_i$ updates its local model age (\autoref{alg:cs}, l.~\ref{l:age_update}). As the aggregation of a server model embeds more than one client update, we apply a weighting strategy to increase a model's age after aggregating another server's model.

\section{Performance Evaluation}
\label{sec:performance_evaluation}

\begin{figure*}
\centering
\begin{minipage}{\columnwidth}
  \centering
        \includegraphics[width=\columnwidth]{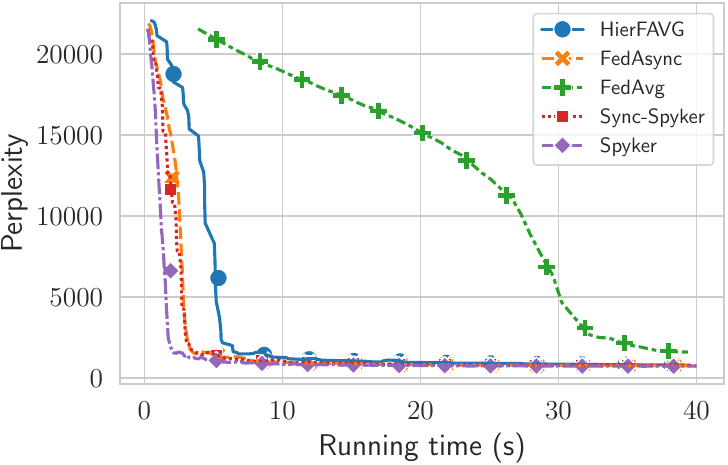}
        \captionof{figure}{WikiText2: Perplexity wrt. time. (lower is better)}
        \label{fig:wiki_time}
\end{minipage}
\hfill
\begin{minipage}{\columnwidth}
  \centering
        \includegraphics[width=\columnwidth]{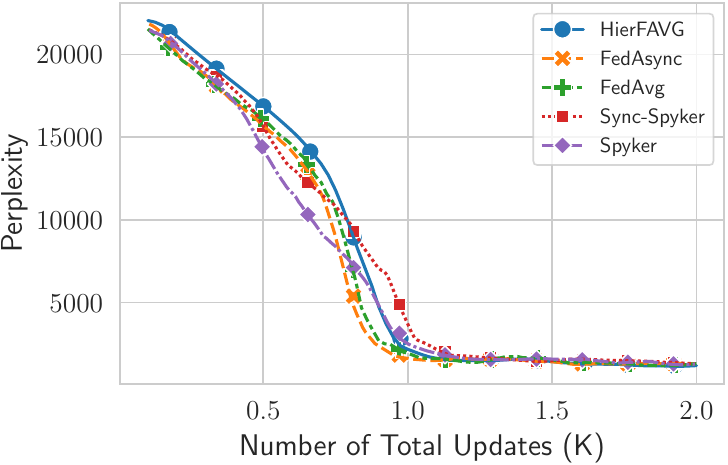}
        \captionof{figure}{WikiText2: Perplexity wrt. \# updates (lower is better)}
        \label{fig:wiki_updates}
\end{minipage}
\end{figure*}

In this section, we describe our experimental settings, and compare \ma to state-of-the-art representative FL algorithms. We show the superior accuracy of \ma over previous baselines, demonstrate its improved scalability with the number of servers, and report the positive impact of our learning rate decay function on accuracy. We also report the network consumption of all algorithms.   

\subsection{Settings}

\textbf{Baselines.} We compare \ma with three recent Federated Learning algorithms: FedAvg~\cite{fl_proposed}, FedAsync~\cite{xie2019fedasync} and HierFedAvg~\cite{client_edge_cloud}. FedAvg is the original synchronous FL framework, while FedAsync and HierFedAvg are respectively state-of-the-art asynchronous and hierarchical FL frameworks.
We also implemented a synchronous version of \ma, which we call \ms, that uses a synchronous server-server model aggregation procedure. \ms periodically triggers synchronous model exchanges between servers (i.e., ignoring the influence of $h_{inter}$ to trigger model exchanges).  
Periodically, the servers broadcast their model, along with their age, and wait for all other server models. Upon receiving all models, a server aggregates them following a deterministic order (i.e., using the server IDs).  
Following this exchange and aggregation of server models, all servers own the same model.
In \ms, when the server synchronization algorithm is started, servers stop processing local updates from clients, and instead store them. Client updates are then processed when all the server models have arrived and have been aggregated. \\

\textbf{Datasets.} We conduct experiments on the MNIST and CIFAR-10 image collections, and on the WikiText2 language modeling dataset. For a given experiment, a dataset is split into subsets of equal sizes that are assigned to different clients. To introduce client data heterogeneity, we assign $l$ labels to each client. A smaller $l$ indicates a higher degree of  non-\gls*{iid} data distribution and thus larger heterogeneity among clients. We set $l$ to 2 in non-\gls*{iid} data experiments. \\

\textbf{Computation and Network Delays.}
\label{subsec:experiment_settings}
Our experiments take into account the computations that clients execute. Time is maintained at each client, and is advanced depending on the procedure it executes during the experiments. \autoref{tab:computing_delays} details the average computing times of the server aggregation procedure of all FL frameworks. These values were obtained by benchmarking all algorithms using the Python $time$ package. To simulate client heterogeneity, we sample each client's training delay from a Gaussian distribution $N(\mu, \sigma^2)$, where $\mu = 150$ and $\sigma=7.5$. 
We use the same training delays per client across all experiments.
Since servers might be located in different regions of the world (e.g., Amsterdam and Sydney), we set the communication delays among servers according to the Amazon Web Service network latency~\cite{awsLatency} as shown in \autoref{tab:delays}. We assume that a client and its nearest server are in the same geographical location and therefore set their communication delay using the diagonal of~\autoref{tab:delays}. We assume that all network links have a 100\,Mbps bandwidth. For each experiment, we detail the number of servers and clients that we use along with the parameter values. \\

\begin{table}[t]
    \centering
    \caption{Computation time required per procedure (ms).}
    \normalsize
    \begin{tabular}{|r|c|}
        \hline
             Local Training  &  200 \\
             Model Aggregation in \ms &  2 \\
             Model Aggregation in \ma &  2 \\
             Model Aggregation in FedAVg &  15 \\
             Model Aggregation in HierFAVG  & 15 \\
             Model Aggregation in FedAsync &  2 \\
        \hline
    \end{tabular}
    
    \label{tab:computing_delays}
\end{table}

\begin{table}[t]
    \centering
    \caption{Communication delays between geographical locations (ms).}
    \normalsize
    \begin{tabular}{|c|c|c|c|c|}
    
         \cline{2-5}
         \multicolumn{1}{c|}{} & Hongkong & Paris & Sydney & California\\
         \hline
         Hongkong  & 1.41 & 194.9 & 132.28 & 155.13 \\
         \hline
         Paris  & 197.91 & 0.9 & 278.83 & 142.25 \\
         \hline
         Sydney  & 132.06 & 280.11 & 2.56 & 138.47\\
         \hline
         California  & 154.96 & 142.79 & 138.57 & 2.14\\
    \hline
    \end{tabular}
    
    \label{tab:delays}
\end{table}

\begin{figure*}
\centering
\begin{minipage}{\columnwidth}
  \centering
  \includegraphics[width=\columnwidth]{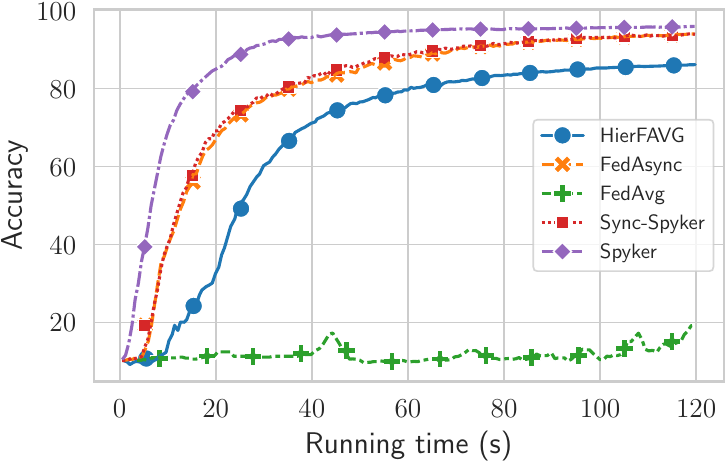}
  \captionof{figure}{MNIST: Accuracy wrt. time (higher is better)}
  \label{fig:mnist_time}
\end{minipage}
\hfill
\begin{minipage}{\columnwidth}
  \centering
  \includegraphics[width=\columnwidth]{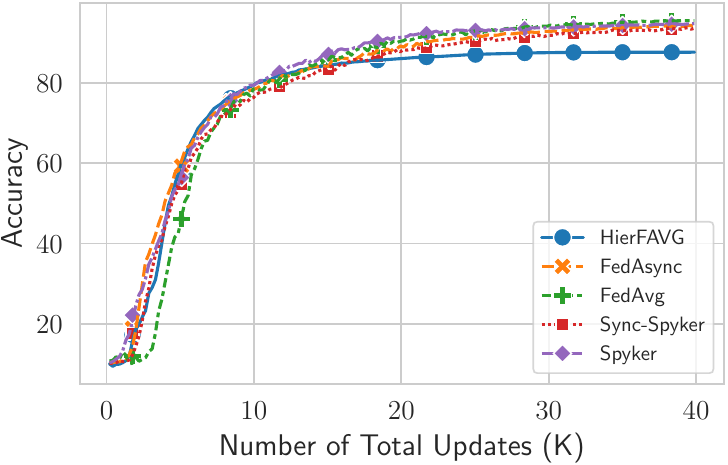}
  \captionof{figure}{MNIST: Accuracy wrt. \# updates (higher is better)}
  \label{fig:mnist_updates}
\end{minipage}
\end{figure*}

\begin{figure*}
\centering
\begin{minipage}{\columnwidth}
  \centering
  \includegraphics[width=\columnwidth]{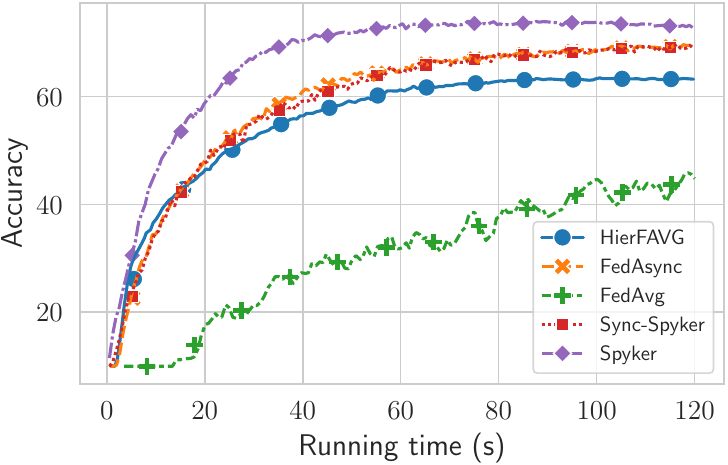}
  \captionof{figure}{CIFAR-10: Accuracy wrt. time (higher is better)}
  \label{fig:cifar_time}
\end{minipage}
\hfill
\begin{minipage}{\columnwidth}
  \centering
  \includegraphics[width=\columnwidth]{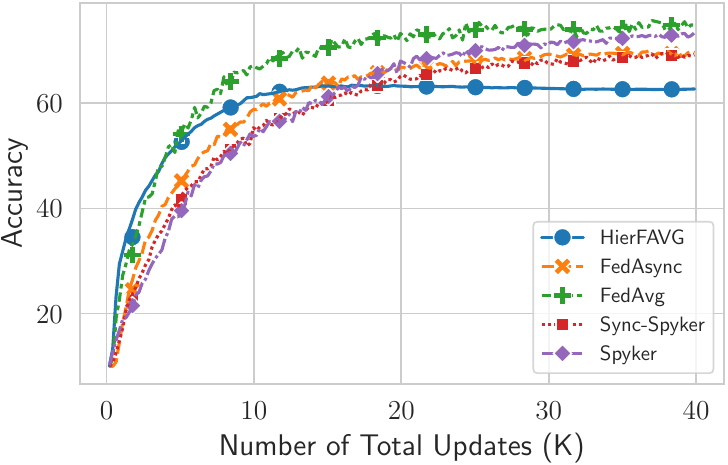}
  \captionof{figure}{CIFAR-10: Accuracy wrt. \# updates (higher is better)}
  \label{fig:cifar_updates}
\end{minipage}
\end{figure*}

\textbf{Model.}  For MNIST, we use a CNN model with two convolutional and two fully connected layers. For CIFAR-10, we use a CNN model with three convolutional and two fully connected layers. For WikiText2, we use the next character \gls*{lstm} neural network model, which has been designed for character-level text generation tasks. The language model consists of an embedding layer that is used to capture the semantic and syntactic properties of characters, an \gls*{lstm} layer that captures dependencies between characters and generates coherent text, and a fully connected layer that transforms the \gls*{lstm}'s hidden states into a probability distribution over all possible characters in the vocabulary. The initial local learning rate of clients $\eta_k$ is $0.05$. In asynchronous settings, we use $\alpha=0.5$ for the staleness weighting in FedAsync and a global learning rate of $\eta=0.6$ for the client-server update. 

\subsection{Accuracy}
\label{subsec:comparison_of_5_algorithms}

For the following experiments, we use $100$ clients that are equally distributed across $4$ servers located in the AWS region we consider. \autoref{fig:wiki_time}, \autoref{fig:mnist_time} and \autoref{fig:cifar_time} show that \ma converges faster than other baselines in terms of elapsed time for all three datasets. \ma and \ms converge as quickly as other baselines in terms of number of processed client updates. 

\begin{table}[t]
    \centering
    \caption{Multiplicative factor for the amount of time and number of updates necessary for algorithms to reach a $90\%$ accuracy with $200$ and $300$ clients compared with their own execution with $100$ clients. A low value indicates that an algorithm scales well with the number of clients. }
    \label{tab:ntest}
    \normalsize
    \begin{tabular}{|c|c|c|c|c|}
        
        \cline{2-5}
        \multicolumn{1}{c}{}  & \multicolumn{2}{|c|}{200 clients} & \multicolumn{2}{c|}{300 clients }\\ 
        
        \cline{2-5}
        \multicolumn{1}{c}{} & \multicolumn{1}{|c|}{Time} & Updates & Time & Updates\\ \hline 
        FedAsync & 2.42 &  3.63 & 5.99 & 9.01\\ \hline
        FedAvg & 1.98 & 4.05  & 2.44 & 7.26 \\ \hline
        HierFAVG & 1.64 & 3.18 & 1.75 & 8.19\\ \hline
        \hline
        \ma & $\mathbf{1.21}$ & $\mathbf{2.43}$ & $\mathbf{1.43}$ & $\mathbf{4.34}$ \\ \hline
        \ms & 1.61 & 2.68 & 1.90 & 4.87\\ \hline   
    \end{tabular}
\end{table}

We also evaluated the ability of the $5$ FL frameworks to scale with the number of clients.  
We evaluated the time each algorithm needs to reach $90\%$ accuracy with MNIST and $4$ servers using $100$, $200$ and $300$ clients. \autoref{tab:ntest} reports this time for all algorithms for $200$ and $300$ clients divided by the time measured with $100$ clients. 
When the number of clients in the system increases, the convergence time and required number of updates of the various methods increase following different trends. \ma is the method that scales the best. For example, it requires $21\%$ more time to converge with $200$ clients than with $100$ clients while other baselines (excluding \ms) require at least $64\%$ more time. It also  requires $2.43$ times as many updates to converge while other baselines require at least $3.18$ times as many. 
On the other side of the spectrum, FedAsync's required time and number of updates to converge increase the most. With $200$ clients, its convergence time is multiplied by $2.42$, while its required number of updates is multiplied by $3.63$

\ma not only exhibits the fastest convergence, but it also scales the best with the number of clients. 

\autoref{fig:wiki_updates}, \autoref{fig:mnist_updates} and \autoref{fig:cifar_updates} show the perplexity of all the baselines, \ma{} and \ms{} on WikiText-2, and their accuracy on MNIST and CIFAR-10 depending on the number of client updates processed. One can see that \ma{} and \ms{} do not always require the least number of updates to reach a given perplexity or accuracy. This is not a problem in practice as it makes more sense to evaluate the evolution of perplexity or accuracy based on elapsed time during a deployment. However, these figures also show that \ma{} reaches similar perplexity or accuracy levels as FedAvg, which generally converges the fastest based on the number of updates (but not based on elapsed time, where it is the worst performing baseline).     

\begin{figure*}
\centering
\begin{minipage}{\columnwidth}
  \centering
    \includegraphics[width=\linewidth]{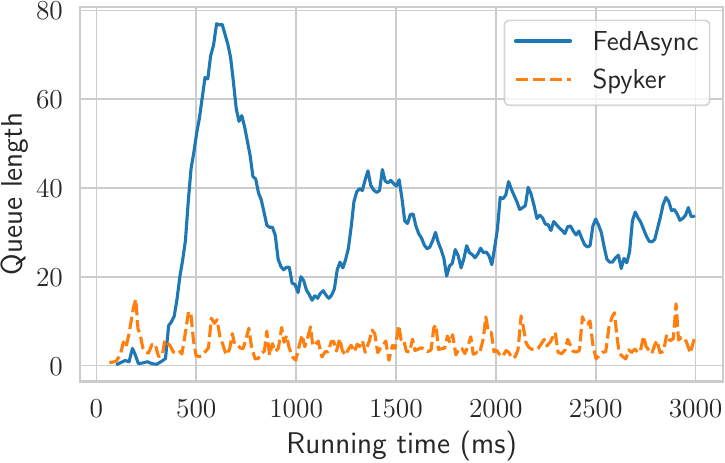}
    \captionof{figure}{Number of updates queued at a server with \ma and FedAsync}
    \label{fig:qlen}
\end{minipage}
\hfill
\begin{minipage}{\columnwidth}
  \centering
    \includegraphics[width=\linewidth]{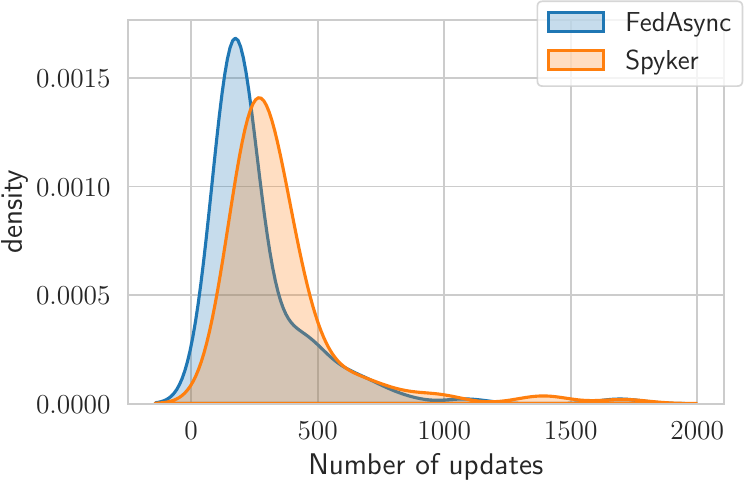}
    \captionof{figure}{Density distribution of the number of updates per client with \ma and FedAsync}
    \label{fig:kde}
\end{minipage}
\end{figure*}

\subsection{Impact of Multiple Servers in Asynchronous FL}
\label{sec:multi_server_afl}

In asynchronous FL systems, the server aggregates an update immediately after receiving it. However, a busy server can cause a queueing of updates that await being processed. 
To further illustrate the benefits of introducing a multi-server system to an asynchronous FL framework, we evaluate the update queuing phenomenon with \ma ($4$ servers) and FedAsync ($1$ server)  with $200$ clients. The local training delays of clients are generated using a Gaussian distribution with a mean of $150$\,ms and a standard deviation of $60$\,ms. Processing delays are set using \autoref{tab:delays}.  

\autoref{fig:qlen} shows that FedAsync's queue length increases to nearly $80$ updates from $300$\,ms to $600$\,ms and always stays above $20$ during the rest of the experiment. 
Client updates are processed more efficiently by the $4$ servers that \ma uses, as server queues never contain more than $20$ updates. 

We further conducted a group of  experiments to evaluate the time needed for FedAsync and \ma to reach $90\%$ and $95\%$ accuracy on the MNIST dataset. For these experiments, the time difference observed is only caused by resource heterogeneity as we set all network latency's to the same value. The results are shown in the bottom three rows of \autoref{tab:ma_boost}. \ma reaches a $90\%$ and a $95\%$ accuracy respectively $38\%$ and $25\%$ faster than FedAsync. 

\begin{table}[t]
\centering
\caption{   
    Time to reach $90\%$ or $95\%$ accuracy with FedAsync and \ma.     
    \textit{Lat} uses AWS latency~\cite{awsLatency}, and \textit{No lat} assumes uniform latency between servers (equal average in both cases).  
}
    \normalsize
    \begin{tabular}{|c|c|c c|}
        \hline
        Network & Method & Time 90\% & Time 95\%\\ \hline
        \multirow{3}{*}{Lat.}  
            & FedAsync & $59s$ & $125s$ \\ 
            & \ma & $22s$ & $51s$ \\ \cline{2-4}
            & Improvement & $\textbf{-61\%}$ & $\textbf{-58\%}$ \\ \hline
        \multirow{3}{*}{No lat.}  
            & FedAsync & $40s$ & $75s$ \\
            & \ma & $25s$ & $56s$ \\ \cline{2-4}
            & Improvement & $\textbf{-38\%}$ & $\textbf{-25\%}$ \\ \hline
    \end{tabular}
\label{tab:ma_boost}
\end{table}

To demonstrate that \ma also  mitigates the impact of high communication delays, we conducted a similar group of experiments that, in addition, introduce network latencies (cf. \autoref{tab:delays}). The results are shown in the top three rows of \autoref{tab:ma_boost}.
Compared with the bottom three rows, the top three rows show a larger difference between \ma and FedAsync. \ma appears $61\%$ faster to reach $90\%$ accuracy, and $58\%$ faster to reach $95\%$ accuracy than FedAsync. 
This difference can be explained by the fact that \ma decreases the distance between servers and clients compared to FedAsync, which uses a single server. 

\begin{figure*}
\centering
\begin{minipage}{\columnwidth}
  \centering
    \includegraphics[width=\columnwidth]{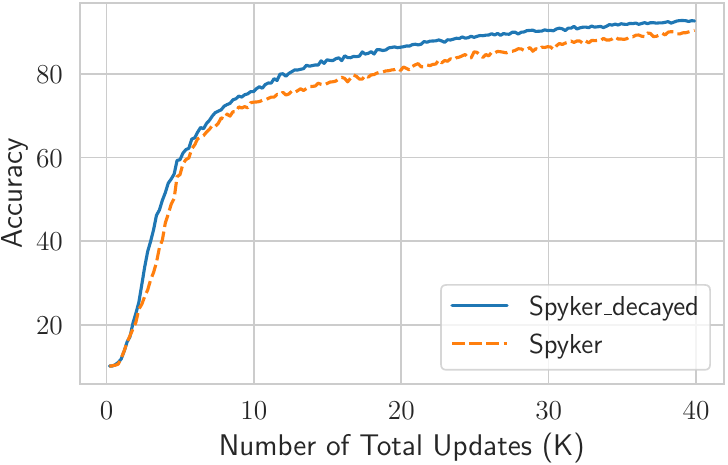}
    \captionof{figure}{Accuracy with and without a learning rate decay on MNIST with $4$ servers and $100$ clients ($25$ per server)}
    \label{fig:decayed}
\end{minipage}
\hfill
\begin{minipage}{\columnwidth}
  \centering
    \includegraphics[width=\linewidth]{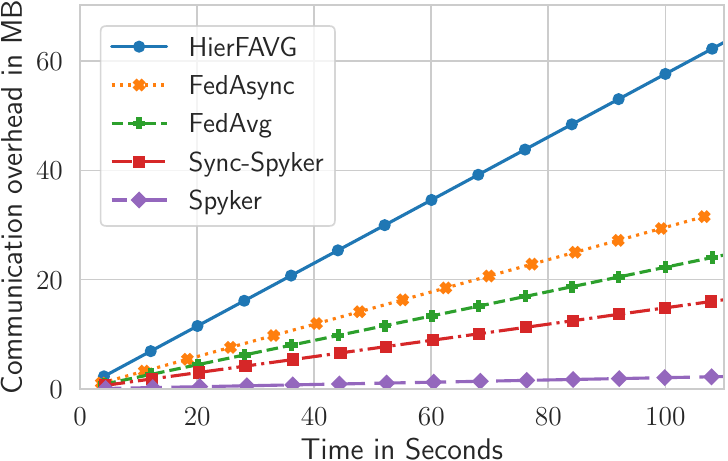}
    \captionof{figure}{Overall network consumption of \ma compared to the baselines}
    \label{fig:communication_overhead}
\end{minipage}
\end{figure*}

\begin{table}[t]
    \centering
    \caption{Effect of imbalanced number of clients per server on accuracy. }
    \label{tab:clients_per_server}
    \normalsize
    \begin{tabular}{|c|c|c|c|c|}
        
        \cline{2-5}
        \multicolumn{1}{c}{}  & \multicolumn{4}{|c|}{\ma}  \\

        \cline{2-5}
        \multicolumn{1}{c}{} & \multicolumn{1}{|c|}{25 clients} & 52 clients & 63 clients & 70 clients\\ \hline 
        Accuracy & $96.4\%$ &  $-14.5\%$ & $-11.9$&  $-7.63\%$\\ \hline
        Duration (s) & 120 & $+17.76$  & $+33.04$ & $+53.08$ \\ \hline

    \end{tabular}
\end{table}

\subsection{Number of Clients per Server}
\label{subsec:n_clients_per_server}

We evaluate the effect of imbalanced number of clients allocated to each server.
We consider four different scenarios. The first one consists of 4 servers with each 25 clients. In the three other scenarios, one of the servers has more clients (i.e., it has 52, 63, or 70 clients) while the remaining clients are divided evenly over the remaining servers. Table~\ref{tab:clients_per_server} show the accuracy of the global model and the time necessary for it to converge. These results show that slight imbalance in the client distribution degrades accuracy the most (-14.5\% with the second most balanced scenario) and that increasing imbalance always increases the convergence time. In comparison, we observed that HierFAVG's performance is less impacted by this kind of imbalance, but its performance nonetheless is always worse than \ma{}'s.  

\subsection{Impact of Learning Rate Decay}
\label{subsec:impact_of_learning_rate_decay}

To demonstrate the positive impact of our learning rate decay method on \ma, we conducted two experiments. 
We first used \ma and FedAvg each with $100$ clients to perform the classification task on the MNIST dataset. 
For these two algorithms that represent the single-server and multi-server baselines, we gathered statistics on the number of updates of every client and plotted a \gls*{kde} plot to show its distribution. 
The most desirable \gls*{kde} plot for a FL system should exhibit a single concentrated peak, which would mean that clients contribute equally. \autoref{fig:kde} shows the \gls*{kde} plots of \ma and FedAsync.

FedAsync has a more centralized distribution of updates with a steep peak at around $200$ updates compared to a gentler peak for \ma at around $300$ updates. The reason for this difference lies in the fact that the major source of delay in FedAsync is communication, and the range of possible  communication delays with clients is wider with a single server. On the other hand, by introducing multiple servers, the range of possible communication delays with \ma becomes smaller and the  differences between client training times becomes the major source of heterogeneity. 
We also observe a peak at around $1,400$ updates for \ma and $1,700$ for FedAsync, which comes from the presence of fast  clients near the server, which could cause a biased global model due to their higher influence on the server's model if left unaddressed. 
This analysis justifies the need for a client learning rate decay mechanism that customizes the learning rate of clients according to the number of updates they contribute. The impact of the updates that the most active clients generate is therefore dampened to balance the overall contribution of clients. Interestingly however, the decay function would use a smaller decay ratio if the system as a whole would catch up with the fast clients. 

In the second experiment, we evaluated \ma's convergence speed with and without our learning rate decay mechanism.  \autoref{fig:decayed} shows that introducing the learning rate decay mechanism increases \ma's convergence speed, and therefore that it reduces the impact of client data heterogeneity.

\subsection{Bandwidth consumption}

We evaluated the bandwidth consumption of all FL algorithms over time. We measured the number of bytes transferred over $110$\,s during server-server and server-client model communications. \autoref{fig:communication_overhead} details the network consumption of all algorithms with the MNIST dataset, $4$ servers and $100$ clients equally distributed over all servers.  
FedAvg has the lowest bandwidth consumption with $2.28$\,MB consumed, which can be explained by its synchronous communications and its use of a single server. \ma has the highest bandwidth requirements with $63.4$\,MB transferred because of its asynchronous server-server and server-client communications. This network consumption is however very reasonable for modern networks. For each client update, the aggregated global model is relayed back to the client, which results in higher bandwidth requirements. Similarly, \ms is the second most network intensive algorithm and transmits $32.5$\,MB over the network since it maintains asynchronous client-server interactions. Unlike \ma's asynchronous multi-server aggregation, \ms's synchronous multi-server aggregation halts the client updates, which results in lower communication overhead. FedAsync's asynchronous client updates incur $24.5$\,MB of data transmissions over the network, which is higher than with FedAVG. HierFAVG's synchronous client-server communication results in lower communication overhead of $16.35$\,MB but the hierarchical server aggregation requires relatively higher bandwidth.

\section{Related Work}
\label{sec:related_work}

This section discusses additional related work on asynchronous and multi-server FL that complements~\autoref{sec:background}.

\subsection{Asynchronous FL}

ASO-Fed~\cite{asyn_fl} is an asynchronous FL framework that enables wait-free computation and communication. ASO-Fed allows client updates to stream into the federator in different rounds and each local training is based on the newly streamed-in data.
FedAsync~\cite{xie2019fedasync} is an asynchronous optimization algorithm for FL that guarantees a near-linear convergence and address the staleness problem in the asynchronous setting. Each time an update arrives, the weighted averaging is applied for the current global model and the update.
AsyncFedED is an adaptive asynchronous FL aggregation based on Euclidean distance~\cite{wang2022asyn}. If the global model is updated before a client sends back its update, the learning rate of the server should be adjusted based on the staleness of the current update. AsyncFedED evaluates the staleness of an update by its Euclidean distance from the server.  Larger staleness leads to a smaller learning rate for its aggregation. The staleness of weight is the relative weight differences.  
FedAsync~\cite{xie2019fedasync} introduces a staleness strategy where the influence of a client's update decreases when its staleness grows as the inverse function of $\alpha$, the adaptive parameter.
A larger $\alpha$ indicates the influence of stale updates should be reduced. 
FedAsync performs best when the staleness is small, and it can converge faster than FedAvg. 

Overall, asynchronous FL approaches incur a high computational load on the server and therefore do not scale well with the number of clients. Intelligent node selection techniques~\cite{Tea_fed,HaoZZ20, ChenLHL021} aim at alleviating this burden but are often biased towards specific clients.

\subsection{Multi-Server FL}

Existing multi-server methods~\cite{multiairfed, fedprox, HierFedAsync, hier_sync, client_edge_cloud} add edge servers or assign proxy clients to undertake part of the workload. Hierarchical FL approaches have also been described~\cite{hier_sync, client_edge_cloud, HierFedAsync, smartPC}. 
Qu et al. demonstrated the convergence of a multi-server FL algorithm~\cite{Zhe2023Convergence} where servers indirectly synchronize their models through overlapping client sets. Clustering is frequently used in multi-task and unsupervised FL~\cite{cluster0, cluster1, cluster2}. Incorporating clustering algorithms in our framework for potential refinement is future work.

Xie et al. proposed a multi-center FL algorithm, FeSEM~\cite{FeSEM}, to address the challenge of data heterogeneity. The main idea is to cluster the updates and assign them to their closest global model (i.e. center). They showed in experiments that the clusters of local updates characterize clients' data distribution thus models generated from similar data distributions are gathered on the same center and aggregated. 

HierFAVG~\cite{client_edge_cloud} is a client-edge-cloud hierarchical framework, which can be considered as a multi-server version of the FedAvg algorithm. Each edge server employs averaging aggregation for its clients. After every certain number of rounds, the cloud server applies averaging strategy to the edge servers. The introduction of the hierarchical structure reduces the communication burden on servers and improves the capability of the single-server FedAvg algorithm while maintaining the stability of synchronous systems. 

Despite their ability to support a larger number of clients, current multi-server and hierarchical-server algorithms do not directly address the sensitivity to large system heterogeneity of synchronous systems. 

\subsection{Geo-distributed Machine Learning}

Many distributed ML systems target large-scale ML applications~\cite{dean2012large,chilimbi2014project,cui2016geeps}, however, they assume that network communication happens within a datacenter. 
Cano et al.~\cite{cano2016towards} discussed running a machine learning system in geo-distributed settings. They show that one can leverage a communication-efficient algorithm for logistic regression models~\cite{mahajan2013functional} to improve performance. Several works~\cite{jaggi2014communication,zhang2015disco,hsieh2017gaia} focused on the design of communication-efficient mechanisms that leaves the ML algorithms unmodified. These works are orthogonal to our approach, since we design asynchronous multi-server FL algorithms for the geo-distributed settings. 

\section{Conclusion}
\label{sec:conclusion}

We described \ma, the first fully asynchronous multi-server FL algorithm. \ma  focuses on practical scenarios where clients are distributed over the world, and it addresses the performance bottlenecks that naturally appear in presence of weak  clients or heterogeneous networks. \ma scales better with the number of clients than previous works.  \ma relies on an aging mechanisms that servers use to support asynchrony of all exchanges and on a token-based algorithm that servers rely on to synchronize their model exchanges. Our experimental results show that \ma requires less time to converge than three representative algorithms -- FedAsync, HierFAVG, and FedAvg, and is also faster than its partially synchronous variant \ms where servers synchronize themselves using a synchronous algorithm. 
\ma also leverages a learning rate decay method to maintain high accuracy despite client heterogeneity.
Future work includes exploring the possibility of integrating clustering algorithms in \ma to enable servers to group clients based on possible similarities in their data distributions. 

\newpage
\bibliographystyle{ACM-Reference-Format}
\bibliography{biblio}

\end{document}